\documentclass[a4paper,fleqn,review]{cas-sc}
\usepackage{lscape}
\usepackage{pdflscape}
\usepackage{graphicx}
\usepackage{caption}
\usepackage{subcaption}
\usepackage{amsmath}
\usepackage{array} 
\usepackage{booktabs}
\usepackage{multirow}
\usepackage{array}
\usepackage[labelfont=bf]{caption}
\usepackage{graphicx} 
\usepackage{caption}  
\usepackage{pdfpages}
\usepackage{booktabs}
\usepackage{graphicx} 
\usepackage{float} 
\usepackage{amsmath}
\usepackage{tcolorbox}
\usepackage{ragged2e}  % for \justifying
\usepackage{verbatim}  % for verbatim environment (if not already included)

\usepackage[authoryear]{natbib}

\def\tsc#1{\csdef{#1}{\textsc{\lowercase{#1}}\xspace}}
\tsc{WGM}
\tsc{QE}

\begin{document}
\let\WriteBookmarks\relax
\def\floatpagepagefraction{1}
\def\textpagefraction{.001}

\shorttitle{}    

%Confidence-Driven Multi-Model Framework for Medical Question Answering: Leveraging Model Diversity Across Clinical Benchmarks}
%Multi-model Ensemble with Dynamic, Intelligent Confidence
%Confidence-driven Unified Reasoning Ensemble

\title [mode = title]{CURE: Confidence-driven Unified Reasoning Ensemble Framework for Medical Question Answering}
  
\author[1]{Ziad Elshaer}

\cormark[1]

\ead{zelshaer@nu.edu.eg}

\affiliation[1]{organization={School of Information Technology and Computer Science, Nile University},
            addressline={26th of July Corridor}, 
            city={Sheikh Zayed City},
            postcode={12588}, 
            state={Giza},
            country={Egypt}}

\affiliation[2]{organization={Graduate School of Information Science, University of Hyogo},
             city={Kobe},
            postcode={650-0047}, 
            country={Japan}}          
\affiliation[3]{organization={Advanced Medical Engineering Research Institute, University of Hyogo},
             city={Himeji},
            postcode={670-0836}, 
            country={Japan}}

\author[2,3]{Essam A. Rashed}[orcid=0000-0001-6571-9807]
\ead{rashed@gsis.u-hyogo.ac.jp}

\cortext[1]{Corresponding author}

\begin{abstract}
High-performing medical Large Language Models (LLMs) typically require extensive fine-tuning with substantial computational resources, limiting accessibility for resource-constrained healthcare institutions. This study introduces a confidence-driven multi-model framework that leverages model diversity to enhance medical question answering without fine-tuning. Our framework employs a two-stage architecture: a confidence detection module assesses the primary model's certainty, and an adaptive routing mechanism directs low-confidence queries to Helper models with complementary knowledge for collaborative reasoning. We evaluate our approach using Qwen3-30B-A3B-Instruct, Phi-4 14B, and Gemma 2 12B across three medical benchmarks; MedQA, MedMCQA, and PubMedQA. Result demonstrate that our framework achieves competitive performance, with particularly strong results in PubMedQA (95.0\%) and MedMCQA (78.0\%). Ablation studies confirm that confidence-aware routing combined with multi-model collaboration substantially outperforms single-model approaches and uniform reasoning strategies. This work establishes that strategic model collaboration offers a practical, computationally efficient pathway to improve medical AI systems, with significant implications for democratizing access to advanced medical AI in resource-limited settings.
\end{abstract}


\maketitle

\section{Introduction}

%{\color{red}more references to be added}~\cite{yu2024enhancing,ALONSO2024102938,wang2025nc}

The rapid advancement of artificial intelligence in healthcare has ushered in a new era of possibilities for medical practice, education, and research. Large Language Models (LLMs), with their remarkable ability to understand, process, and generate human-like responses to complex queries, represent one of the most promising technological developments in modern medicine~\citep{brown2020language, touvron2023llama}. These sophisticated AI systems have demonstrated unprecedented capabilities in natural language understanding and reasoning, making them particularly valuable for applications requiring comprehensive knowledge synthesis and contextual understanding.

The medical domain presents a unique and critical application area for LLMs, where the potential impact extends far beyond technological innovation to directly influence patient outcomes, clinical decision-making, and healthcare accessibility. Medical knowledge is characterized by its complexity, specificity, and constant evolution, requiring practitioners to synthesize vast amounts of information from diverse sources including clinical guidelines, research literature, diagnostic protocols, and treatment recommendations~\citep{esteva2019guide, rajkomar2018scalable}. The ability of AI systems to assist healthcare professionals in navigating this complex knowledge landscape has significant implications for improving diagnostic accuracy, reducing medical errors, and democratizing access to expert-level medical knowledge, particularly in underserved regions where specialist expertise may be limited.

The social impact of effective medical AI systems extends beyond individual patient care to encompass broader healthcare system challenges including clinician burnout, healthcare costs, and the growing demand for medical services in aging populations worldwide~\citep{thirunavukarasu2023large}. Furthermore, the COVID-19 pandemic has highlighted the critical need for scalable intelligent systems that can support rapid knowledge dissemination and clinical decision-making during health emergencies and unexpected pandemics. The development of reliable, affordable, and accurate medical AI systems represents not just a technological advancement, but a fundamental tool for addressing global healthcare challenges and improving health outcomes across diverse populations.

Current research in medical AI has established a substantial foundation through rigorous benchmarking efforts utilizing standardized datasets. The medical AI community has converged on three primary benchmarks that collectively assess different aspects of medical knowledge and reasoning capabilities. MedQA, featuring USMLE-style multiple-choice questions, serves as a comprehensive test of clinical reasoning and general medical knowledge, mimicking the challenges faced by medical professionals in licensing examinations~\citep{jin2021medqa}. MedMCQA, comprising over 194K questions across 21 medical subjects derived from Indian medical entrance examinations, provides extensive coverage of healthcare topics and represents one of the most comprehensive medical knowledge assessments available~\citep{pal2022medmcqa}. PubMedQA focuses specifically on biomedical research question answering, requiring models to process and reason over scientific literature with yes/no/maybe responses based on research abstracts, thus testing analytical comprehension of evolving medical research~\citep{jin2019pubmedqa}.

Recent peer-reviewed studies published in prestigious venues such as \emph{nature} and \emph{nature medicine} have documented remarkable progress in medical LLM performance~\citep{singhal2023med,ren2025healthcare}. Med-PaLM 2 achieved 86.5\% accuracy on MedQA, representing a substantial improvement over previous approaches, while recent models have pushed performance boundaries even further. Similarly, PubMedQA has witnessed dramatic improvements, with performance evolving from early baselines to current state-of-the-art results. These advances demonstrate the rapid maturation of medical AI systems and their increasing potential for real-world clinical applications. However, despite these impressive achievements, several critical limitations persist in current approaches to medical AI optimization. Most high-performing medical LLMs rely on extensive fine-tuning procedures that require substantial computational resources, large-scale medical datasets, and significant training time~\citep{chen2023meditron, li2023chatdoctor}. These resource-intensive approaches limit accessibility for smaller healthcare institutions, research organizations, and developing countries that could benefit significantly from advanced medical AI capabilities. Additionally, fine-tuning approaches often lack flexibility, requiring complete retraining when adapting to new domains or incorporating updated medical knowledge.

Recent research has explored various approaches to enhance LLM performance in medical domains, including specialized fine-tuning, retrieval-augmented generation (RAG), and multi-agent frameworks~\citep{xiong2024doctorglm, wang2023huatuogpt}. However, these approaches often require significant computational resources, external knowledge bases, or complex infrastructure. An emerging paradigm in the literature suggests that leveraging \emph{model diversity} combining multiple language models trained on different corpora with distinct knowledge distributions can compensate for individual model limitations without requiring fine-tuning or external knowledge retrieval~\citep{jiang2023llm, wang2024mixture}. This approach is grounded in the observation that different pre-training datasets naturally lead to complementary knowledge bases, where one model's weaknesses may be another's strengths.

Recent studies have demonstrated the potential of ensemble and collaborative approaches in various domains~\citep{shnitzer2023large, wang2023self}. However, these approaches typically employ static architectures that process all queries uniformly, regardless of difficulty or the primary model's confidence level. This uniform processing paradigm does not optimize computational efficiency and does not account for the varying distribution of knowledge across different medical subdomains. Furthermore, existing multi-model approaches lack mechanisms to identify when auxiliary models should be consulted, leading to unnecessary computational overhead for straightforward questions while potentially underutilizing collaborative reasoning for challenging queries.

A critical gap exists in understanding how to effectively orchestrate multiple language models for medical question answering based on \emph{confidence-aware routing mechanisms}. No existing research has developed adaptive frameworks that can intelligently assess a primary model's confidence and dynamically recruit auxiliary models with complementary knowledge bases to address knowledge gaps in real-time. This limitation is particularly significant in medical applications where question complexity varies dramatically, from straightforward factual recall to complex multi-hop reasoning requiring integration of diverse medical concepts, and where no single model possesses comprehensive knowledge across all medical subdomains~\citep{nori2023capabilities}. The absence of confidence-driven, multi-model collaborative architectures represents a substantial barrier to optimizing medical AI systems in resource-constrained settings. Current approaches either rely on computationally expensive fine-tuning or fail to leverage the complementary strengths of models trained on diverse corpora. Static ensemble methods that process all questions uniformly waste computational resources on straightforward queries while potentially missing opportunities for collaborative reasoning on challenging questions. This one-size-fits-all approach limits the potential for achieving optimal performance across different medical scenarios and question types while maintaining computational efficiency.

This study aims to develop and evaluate a confidence-driven multi-model collaborative framework for medical question answering that intelligently leverages model diversity without requiring fine-tuning. Our primary objective is to create an adaptive pipeline that: 
\begin{enumerate}
    \item Assesses the primary model's confidence in answering medical questions, 
    \item Dynamically routes low-confidence questions to helper models with complementary training backgrounds, and
    \item Synthesizes diverse model outputs through structured reasoning to generate accurate final answers.
\end{enumerate}
We seek to demonstrate that strategic model collaboration based on confidence-aware routing can achieve competitive performance across established medical benchmarks (MedQA, MedMCQA, PubMedQA) while maintaining computational efficiency and avoiding resource-intensive fine-tuning procedures. Our contributions are threefold: First, we introduce a novel confidence detection mechanism that enables adaptive routing of medical questions based on the primary model's self-assessed certainty. Second, we demonstrate that leveraging model diversity through strategic collaboration of models trained on different corpora (Qwen3-30B-A3B, Phi-4 14B, Gemma 2 12B) effectively compensates for knowledge gaps without fine-tuning.

\section{Materials and Methods}

This study employs a systematic approach to evaluate and enhance medical question-answering capabilities through a multi-model pipeline. The proposed Confidence-driven Unified Reasoning Ensemble (CURE) framework comprises three key components: carefully selected benchmark datasets, strategically chosen LLMs, and a collaborative inference framework. This section details rationale for dataset and model selection, establishing the foundation for the experimental design.

\subsection{Medical Q/A Datasets}

Three widely recognized medical question-answering benchmark datasets were utilized: MedMCQA~\citep{pal2022medmcqa}, MedQA~\citep{jin2021disease}, and PubMedQA~\citep{jin2019pubmedqa}. These datasets were specifically chosen for their established role as gold-standard benchmarks in evaluating medical AI systems, each offering distinct characteristics that comprehensively assess model performance across various medical reasoning tasks. MedMCQA is a large-scale, multiple-choice question answering dataset compiled from Indian medical entrance examinations (AIIMS and NEET PG). This dataset encompasses questions from 2,400 healthcare topics across 21 medical subjects, including anatomy, pharmacology, and surgery. The questions are designed to test complex medical reasoning and clinical decision-making skills, making them particularly valuable for assessing a model's ability to handle real-world medical scenarios. The multiple-choice format with four options per question provides a robust framework for evaluating both factual recall and analytical reasoning. MedQA serves as our second benchmark, derived from professional medical board examinations across multiple countries, including the United States Medical Licensing Examination (USMLE). This dataset features complex, multi-hop reasoning questions that often include clinical vignettes requiring integration of multiple medical concepts. The questions typically present patient scenarios with detailed clinical information, demanding sophisticated diagnostic reasoning and medical knowledge application. MedQA's multiple-choice structure and emphasis on clinical reasoning make it an excellent complement to MedMCQA. PubMedQA stands apart from the other two datasets through its binary (yes/no) answer format and its strong grounding in biomedical research literature. Questions in PubMedQA are derived from PubMed abstracts, focusing on research-oriented medical queries rather than clinical scenarios. This dataset evaluates a model's ability to comprehend and reason about biomedical research findings, representing a different dimension of medical knowledge compared to the clinical focus of MedMCQA and MedQA. The inclusion of PubMedQA ensures our evaluation covers both clinical decision-making and research comprehension capabilities. The three datasets provide comprehensive coverage of medical AI evaluation, spanning clinical reasoning, medical board examination-level knowledge, and biomedical research understanding. Their complementary nature enable wide-scope assessment of model performance across diverse medical contexts and question formats.

\subsection{Language Language Model Architecture}

Three distinct LLMs, each selected for their unique architectural characteristics and training background, were used to evaluate the proposed framework. This multi-model approach highlights diverse knowledge bases and reasoning capabilities within the proposed pipeline

Qwen3-30B-A3B~\citep{qwen3} serves as the primary model in CURE pipeline, forming the backbone of the benchmark evaluations. This model was selected as the core architecture due to its substantial parameter count (30.5 billion total parameters, with 3.3 billion activated per token in its Mixture-of-Experts (MoE) design) and its innovative hybrid reasoning framework, which enables seamless switching between \emph{thinking} mode (for complex logical reasoning, math, and coding) and \emph{non-thinking} mode (for efficient, general-purpose dialogue). This hybrid approach provides enhanced reasoning capabilities necessary for complex medical question-answering tasks, while maintaining computational efficiency. Qwen3-30B-A3B's extensive pre-training on a high-quality corpus of 36 trillion tokens across 119 languages and dialects, incorporating diverse sources like coding, STEM, reasoning, books, multilingual data, and synthetic content, combined with post-training optimizations for instruction-following and agentic tool integration, positions it as a strong baseline for medical reasoning. Its MoE-based architectural efficiency (featuring 48 layers, Grouped Query Attention with 32 query heads and 4 key/value heads, 128 experts, and 8 activated per task) and state-of-the-art performance on general-domain benchmarks (such as surpassing models like QwQ-32B despite 10$\times$ fewer activated parameters) make it an ideal candidate for adaptation to medical applications.

To complement Qwen3-30B-A3B-Instruct, we integrate two additional models with distinct training backgrounds: Phi-4 14B~\citep{phi4} and Gemma 2 12B~\citep{gemma2}. These models were strategically selected because they were trained on different corpora and possess different knowledge distributions compared to Qwen3-30B-A3B. This diversity is crucial for CURE pipeline architecture, as it allows the auxiliary models to provide complementary information and reasoning perspectives on medical questions that may not have been extensively covered in Qwen3-30B-A3B's training data. Phi-4 14B, developed by Microsoft Research, brings specialized reasoning capabilities refined through a focused training methodology that emphasizes logical reasoning and problem-solving. Its 14 billion parameters and distinctive training approach make it particularly effective at providing alternative reasoning pathways for complex medical questions. Gemma 2 12B, developed by Google DeepMind, offers a third perspective with its 12 billion parameters and unique architectural optimizations. Its training on a carefully curated dataset provides complementary medical knowledge that can assist Qwen2.5 in handling edge cases and questions requiring diverse knowledge sources.

\begin{figure}%[htbp]
\centering
\includegraphics[width=\textwidth]{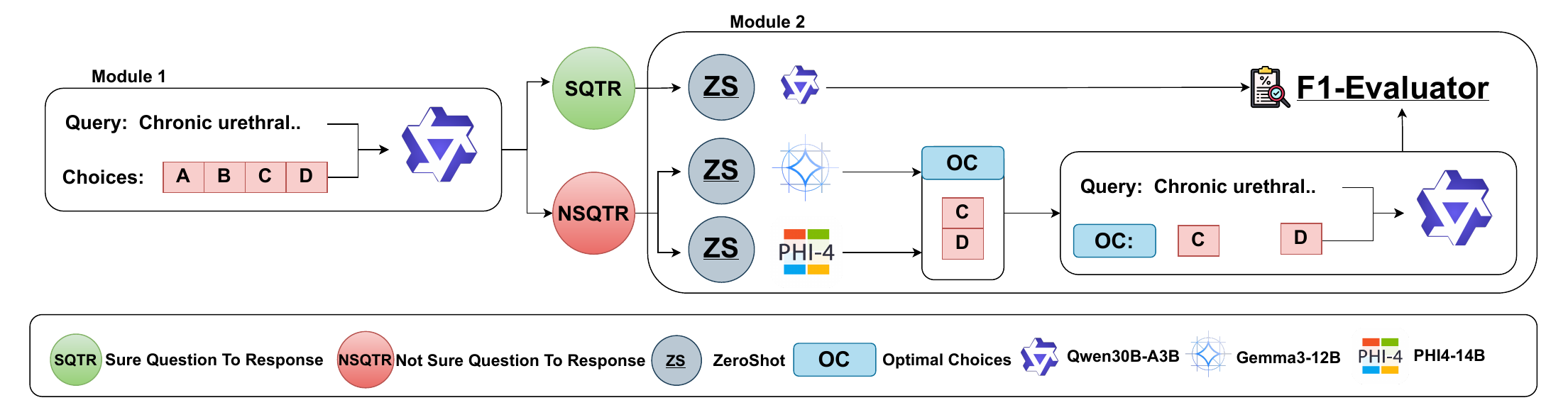}
\caption{Overview of CURE pipeline workflow, showing the confidence detection module and the two processing pathways for high-confidence and low-confidence questions.}
\label{fig:pipeline}
\end{figure}

\subsection{Pipeline Workflow}

The proposed CURE pipeline employs a confidence-driven, multi-stage architecture designed to enhance medical question-answering accuracy by leveraging the complementary strengths of multiple language models. The workflow consists of two primary modules: Module 1: Confidence Detection  and Module 2: Adaptive Answer Generation. This approach enable automaic dynamic route questions based on the primary model's confidence level, ensuring optimal resource utilization while maximizing prediction accuracy. Figure~\ref{fig:pipeline} illustrates the complete CURE pipeline, depicting the flow of questions through both confidence-driven pathways.

\subsubsection{Confidence Detection Module} 

The first stage of CURE pipeline implements a zero-shot confidence detection mechanism. When a question from any of thr benchmark datasets (MedMCQA, MedQA, or PubMedQA) enters the system, it is first processed by Qwen3-30B-A3B-Instruct to assess the model's confidence in providing an accurate answer. This confidence assessment is performed using a zero-shot prompting approach, where the model evaluates whether it possesses sufficient knowledge and certainty to answer the question correctly. The confidence detection serves as a critical decision point that bifurcates our pipeline into two distinct pathways based on the model's self-assessed confidence level.

If Qwen3-30B-A3B-Instruct indicates high confidence in its ability to answer the question, the query is processed directly through the primary model without requiring additional computational resources. In this high-confidence pathway, Qwen3-30B-A3B-Instruct generates its answer and selects the most appropriate option from the available choices. This direct pathway minimizes computational overhead for questions where the primary model demonstrates strong certainty, allowing for efficient processing of straightforward medical queries that fall within the model's well-established knowledge domains. The detailed prompt template used for confidence detection can be found in Appendix~\ref{sec:prompts}.

\subsubsection{Adaptive Answer Generation} 

Conversely, if Qwen3-30B-A3B-Instruct expresses low confidence or uncertainty regarding the question, the query is redirected to the collaborative reasoning module, which considers the diverse knowledge bases of the auxiliary models. This redirection mechanism is crucial for addressing potential knowledge gaps in Qwen3-30B-A3B's training data, as it recognizes that different models trained on distinct corpora may possess complementary strengths in various medical subdomains.

Questions identified as low-confidence are routed through a multi-model reasoning process designed to compensate for these knowledge gaps. In this pathway, the low-confidence question is simultaneously presented to both Phi-4 14B and Gemma 2 12B for independent analysis. Each auxiliary model processes the question using its unique knowledge base and reasoning capabilities, generating its own answer selection from the available options. Since these models were trained on different corpora with varying emphases and knowledge distributions, they can often provide accurate answers to questions that fall outside Qwen3-30B-A3B's strong knowledge domains, effectively filling in the gaps where the primary model lacks confidence.

Following the independent predictions from Phi-4 14B and Gemma 2 12B, their selected answers are aggregated to form a set of candidate optimal choices. These candidate answers represent the collective wisdom of models with different training backgrounds and knowledge distributions, providing diverse perspectives on the correct answer. The aggregated candidates, along with the original question, are then fed back to Qwen3-30B-A3B-Instruct for final reasoning and answer selection, creating a collaborative decision-making process that combines multiple sources of medical knowledge.

This final stage employs a Chain-of-Thought (CoT) prompting template to guide Qwen3-30B-A3B through a structured reasoning process. The CoT prompt explicitly informs the model that the candidate choices provided by Phi-4 and Gemma 2 represent potentially correct answers derived from alternative knowledge sources, encouraging the primary model to carefully consider these suggestions rather than relying solely on its initial uncertain assessment. The prompt guides Qwen3-30B-A3B to review the original question alongside the candidate answers, reason through the medical concepts and knowledge required to answer the question, evaluate each candidate answer based on its medical validity and alignment with the question requirements, and ultimately generate a final answer selection supported by explicit reasoning. The complete CoT prompt template is provided in Appendix~\ref{sec:prompts}.

By incorporating the diverse perspectives of Phi-4 14B and Gemma 2 12B into Qwen3-30B-A3B's decision-making process through structured CoT reasoning, this collaborative approach significantly enhances the accuracy of predictions on challenging questions. The reasoning process ensures that the final answer is not selected arbitrarily but is grounded in logical medical reasoning informed by multiple knowledge sources, each contributing their unique strengths. This confidence-driven, collaborative architecture allows our pipeline to maintain high efficiency on straightforward questions while dedicating additional computational resources and multi-model reasoning to more challenging queries, ultimately improving overall performance across all three benchmark datasets and demonstrating the effectiveness of combining model diversity with adaptive routing strategies.

\section{Results}

In this section, we present the empirical evaluation of CURE framework for medical question answering. The results are organized into three key sections to provide a comprehensive analysis of framework performance. First, we compare CURE zero-shot performance against state-of-the-art benchmarks from recent literature, demonstrating its competitive edge in resource-efficient settings. Second, we detail the framework's performance across the MedQA, MedMCQA and PubMedQA datasets, highlighting dataset-specific strengths and overall robustness. Finally, we conduct an ablation study to isolate the contributions of our pipeline's core components, validating the efficacy of adaptive routing and multi-model collaboration.

\begin{table}%[t!]
\centering
\caption{Accuracy comparison with state-of-the-art approaches MedQA, MedMCQA and PubMedQA datasets. CURE framework achieves competitive results, particularly on PubMedQA, outperforming several benchmarks in multi-step reasoning. Bold indicates leading performance and underline indicated runner-up.}
\label{tab:medical_qa_comparison}
\footnotesize
\begin{tabular*}{\textwidth}{@{\extracolsep{\fill}}lccccl@{}}
\toprule
\multirow{2}{*}{\centering\textbf{Methods}} & \multicolumn{3}{c}{\textbf{Accuracy}} & \multirow{2}{*}{\textbf{Avg Score}} & \multirow{2}{*}{\textbf{Reference}} \\
\cmidrule(lr){2-4}
& \textbf{MedQA} & \textbf{MedMCQA} & \textbf{PubMedQA} & \\
\midrule
\multicolumn{5}{l}{\emph{General/Frontier LLMs }} \\
\midrule
Med-PaLM 2 & \textbf{0.865} & 0.745 & 0.785 & 0.798 & \cite{singhal2023medpalm2}\\
GPT-3.5 & 0.602 & 0.627 & 0.782 & 0.670 & \cite{olson2024}\\
Med-Gemini & 0.820 & 0.760 & 0.890 & \underline{0.823} &\cite{LIEVIN2024100943} \\
LLM-MedQA & 0.810 & \underline{0.770} & 0.840 & 0.807 & \cite{llmmedqa2025}\\
Vicuna-13B & 0.241 & 0.261 & \underline{0.932} & 0.478 & \cite{yang2025large}\\
\midrule
\multicolumn{5}{l}{\emph{Medical LLMs/Frameworks}} \\
\midrule
MedAgentsBench & 0.800 & 0.700 & 0.820 & 0.773 & \cite{medagentsbench2025}\\
OpenBioLLM-Llama3-70B & 0.780 & 0.740 & 0.810 & 0.777 & \cite{openbiollm2024}\\
Me LLaMA & 0.790 & 0.750 & 0.830 & 0.790 & \cite{mellama2025}\\
%MedLLaMa-13B & 0.236 & 0.310 & 0.861 & -- & \cite{yang2025large}\\
%MedAlpaca-13B & 0.373 & 0.380 & 0.953 & -- & \cite{yang2025large}\\
MedXpertQA & \underline{0.830} & 0.720 & 0.870 & 0.807 & \cite{medxpertqa2025}\\
MedQA-CS & 0.775 & 0.735 & 0.795 & 0.768 & \cite{medqacs2024}  \\
Med-PaLM 2 & 0.676 & 0.576 & 0.790 & 0.681 &\cite{singhal2025toward}\\
MEG-LLaMa-8B &   0.660 & 0.606 & 0.780 &0.682&\cite{cabello2024meg}\\
MEG-Mistral-1B & 0.546 & 0.564 & 0.746 &0.619&\cite{cabello2024meg}\\
MEG-Mistral-3B & 0.608 & 0.584 & 0.744 &0.645&\cite{cabello2024meg}\\
\midrule
\multicolumn{5}{l}{\emph{Prompt Engineering Techniques }} \\
\midrule
CoT Ensemble & 0.602 & 0.627 & 0.782 & 0.670 & \cite{olson2024}\\
Interactive CoT & 0.650 & 0.617 & 0.780 & 0.682 & \cite{comparative2025}\\
Self-Reflective CoT & 0.710 & 0.680 & 0.810 & 0.733 & \cite{medagentsbench2025} \\
Explainable LLM & 0.680 & 0.710 & 0.871 & 0.754 & \cite{explainablepubmed2025}\\
\midrule
\textbf{CURE (Proposed)} & 0.741 & \textbf{0.780} & \textbf{0.950} & \textbf{0.824} \\
\bottomrule
\end{tabular*}
\vspace{0.5em}
\normalsize
\end{table}

\subsection{Model Benchmarking}

%To evaluate the effectiveness of our confidence-driven multi-model pipeline, we compare its 
The proposed CURE zero-shot performance is compared against state-of-the-art methods from recent literature, as summarized in Table~\ref{tab:medical_qa_comparison}. Notably, CURE operates entirely in a zero-shot setting without any fine-tuning, retrieval-augmented generation (RAG), or additional enhancements, relying solely on adaptive routing and collaborative CoT reasoning across modestly sized models (Qwen3-30B-A3B-Instruct, Phi-4 14B, and Gemma 2 12B). This lightweight approach yields impressive results, particularly on PubMedQA where it achieves the highest accuracy at 95.0\%, surpassing even specialized techniques like explainable LLMs and multi-agent frameworks. Similarly, on MedMCQA, CURE leads with 78.0\% accuracy, demonstrating superior handling of diverse, multi-subject medical entrance exam questions compared to both general-purpose LLMs (e.g., Med-Gemini at 76.0\%) and prompt-engineered variants (e.g., interactive CoT at 61.7\%).

On MedQA, while larger, domain-specific models like Med-PaLM 2 attain the top score of 86.5\%, our framework's 74.1\% remains competitive given the stark differences in complexity: Med-PaLM 2 involves a 540B-parameter model with extensive medical fine-tuning and instruction-tuning, whereas ours avoids such resource-intensive processes. This gap underscores the efficiency of our adaptive collaboration strategy, which leverages model diversity to bridge knowledge gaps without customization. Across all datasets, our average score of 82.4\% edges out the competition, highlighting the pipeline's robustness in clinical reasoning tasks.

In conclusion, these results affirm that our zero-shot architecture not only matches or exceeds fine-tuned baselines on evidence-heavy (PubMedQA) and broad-domain (MedMCQA) benchmarks but also offers a scalable alternative to heavyweight medical LLMs for MedQA-style diagnostics. By prioritizing confidence-aware routing over model scale, our framework paves the way for deployable, low-overhead medical QA systems in resource-constrained environments.

\subsection{CURE Performance}

The zero-shot Qwen30B-A3B model was evaluated on the three datasets, each with 1,000 samples. The base zero-shot model achieved accuracies of 0.720, 0.765, and 0.948, respectively. Enhancements via PhiZeroShot and GemmaZeroShot, processed through a CoT mechanism to determine correct answers, contributed additional accuracies of 0.264, 0.289, and 0.615 for PhiZeroShot, and 0.232, 0.277, and 0.481 for GemmaZeroShot. The Agent, representing the final CoT-derived output of PhiZeroShot and GemmaZeroShot, added accuracies of 0.285, 0.260, and 0.470, resulting in final accuracies of 0.741, 0.780, and 0.950 for MedQA, MedMCQA, and PubMEDQA, respectively.

% Grouped bar chart
\begin{figure}%[h]
\centering
\includegraphics[width=\textwidth]{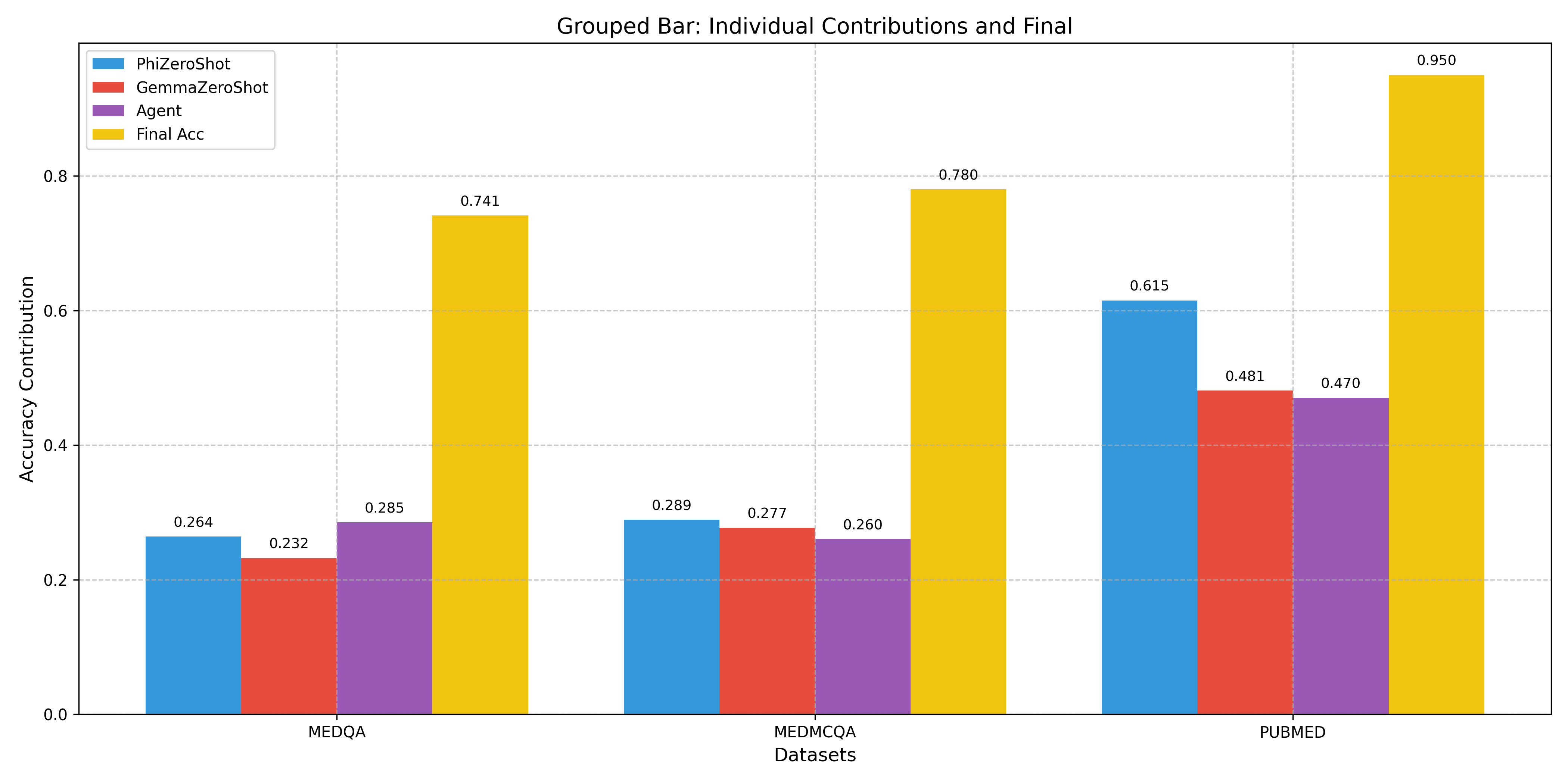}
\caption{Grouped bar chart illustrating the accuracy contributions of PhiZeroShot, GemmaZeroShot, and the Agent (final accuracy) across MedQA, MedMCQA, and PubMedQA datasets. The chart highlights significant enhancements on PubMedQA, with PhiZeroShot and GemmaZeroShot contributing 0.615 and 0.481, respectively, leading to a final accuracy of 0.950.}
\label{fig:grouped_bar}
\end{figure}

% Heatmap
\begin{figure}[h]
\centering
\includegraphics[width=\textwidth]{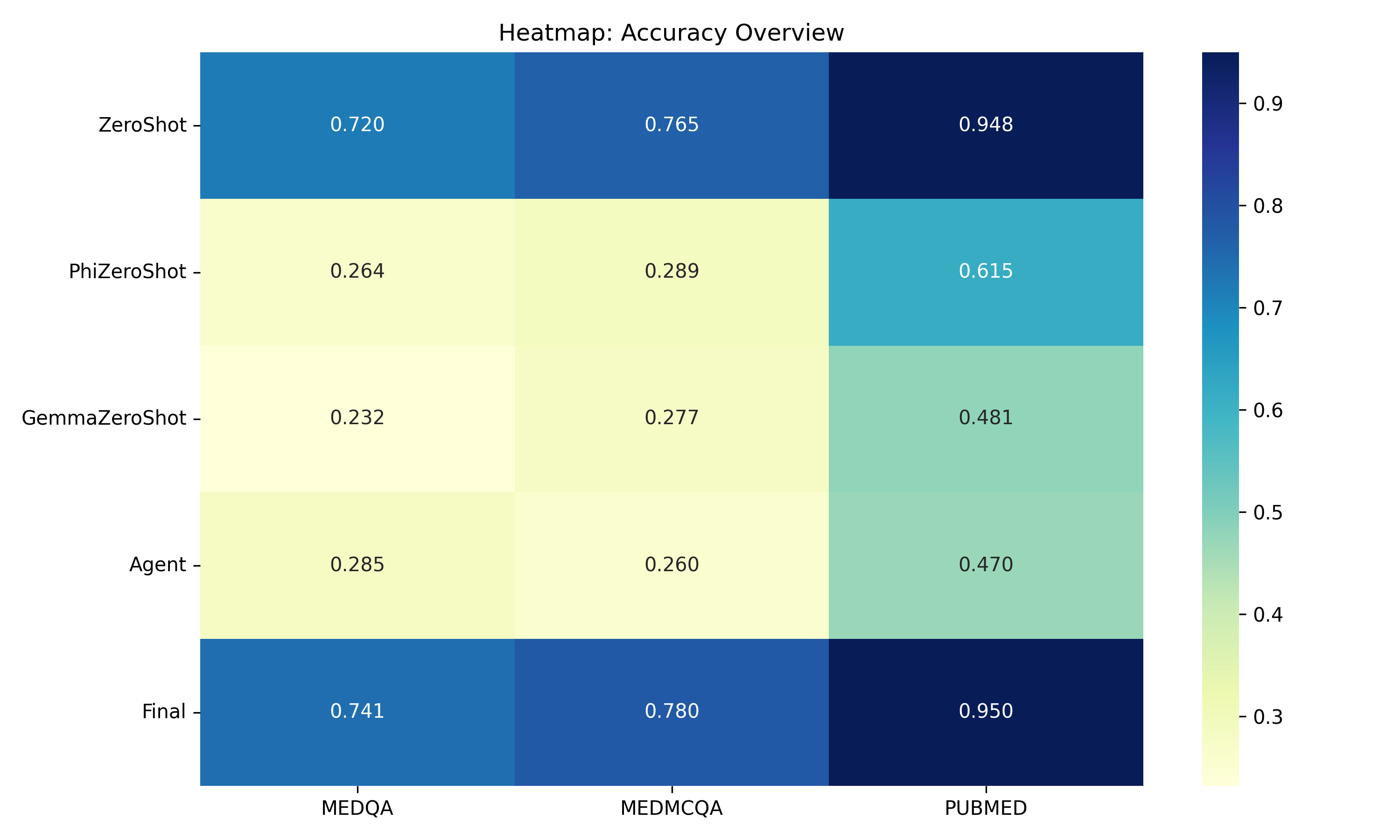}
\caption{Heatmap visualizing the accuracies of ZeroShot, PhiZeroShot, GemmaZeroShot, and Agent (final accuracy) across MEDQA, MEDMCQA, and PUBMED datasets. The heatmap emphasizes the high baseline accuracy of ZeroShot on PUBMED (0.948) and the cumulative enhancements leading to the Agent's final accuracy.}
\label{fig:heatmap}
\end{figure}

% Brief conclusion
The visualizations demonstrate that the CoT-derived enhancements via PhiZeroShot and GemmaZeroShot significantly boost the ZeroShot model's performance, with the Agent achieving the highest final accuracy on PUBMED (0.950).

\begin{table}[t!]
\centering
\caption{Ablation study results showing the impact of pipeline components on accuracy across MedQA, MedMCQA, and PubMedQA datasets. CURE framework demonstrates clear improvements through confidence-aware routing and multi-model collaboration.}
\label{tab:ablation_study}
\footnotesize
\begin{tabular*}{\textwidth}{@{\extracolsep{\fill}}lcccc@{}}
\toprule
\multirow{2}{*}{\centering\textbf{Variant}} & \multicolumn{3}{c}{\textbf{Accuracy}} & \multirow{2}{*}{\textbf{Avg Score}} \\
\cmidrule(lr){2-4}
& \textbf{MedQA} & \textbf{MedMCQA} & \textbf{PubMedQA} & \\
\midrule
Zero-shot Qwen3-30B-A3B & 0.730 & 0.703 & 0.930 & 0.788 \\
CoT Reasoning Qwen3-30B-A3B  & 0.680 & 0.661 & 0.790 & 0.710 \\
\textbf{Full Framework (Confidence Routing + Multi-Model CoT)} & \textbf{0.741} & \textbf{0.780} & \textbf{0.950} & \textbf{0.824} \\
\bottomrule
\end{tabular*}
\vspace{0.5em}
\normalsize
\end{table}

\subsection{Ablation Study}

To understand the contributions of our pipeline's key components: namely, the confidence-driven routing and multi-model collaborative reasoning, we conduct an ablation study comparing simplified variants against the full framework. Table~\ref{tab:ablation_study} presents the accuracy results across the three benchmark datasets.

The baseline zero-shot variant using only Qwen3-30B-A3B achieves solid performance, particularly on PubMedQA at 93.0\%, reflecting the model's strong foundational knowledge in evidence-based biomedical questions. However, applying Chain-of-Thought (CoT) reasoning with the same single model yields mixed results: while it slightly improves PubMedQA to 79.0\% by encouraging structured thinking, it decreases accuracy on MedQA (from 73.0\% to 68.0\%) and MedMCQA (from 70.3\% to 66.1\%), suggesting that solo CoT can introduce reasoning errors or overcomplication for certain clinical and multi-subject queries.

In contrast, the full framework, integrating confidence detection for adaptive routing and collaborative CoT across Qwen3-30B-A3B, Phi-4 14B, and Gemma 2 12B, consistently outperforms both ablations. It boosts MedQA to 74.1\% (a 1.1\% gain over zero-shot and 6.1\% over single-model CoT), MedMCQA to 78.0\% (7.7\% over zero-shot and 11.9\% over single-model CoT), and PubMedQA to 95.0\% (2.0\% over zero-shot and 16.0\% over single-model CoT). The average score rises to 82.4\%, underscoring the value of model diversity in addressing knowledge gaps: low-confidence questions benefit from auxiliary inputs, leading to more robust and accurate final decisions without fine-tuning or external retrieval.

These findings validate our design choices, showing that selective multi-model collaboration enhances zero-shot performance across datasets, especially for challenging queries requiring complementary expertise.

\section{Discussion}

Our confidence-driven multi-model framework achieved competitive performance across three established medical benchmarks without requiring fine-tuning or external knowledge retrieval. The framework demonstrated particularly strong results on PubMedQA (95.0\%) and MedMCQA (78.0\%), with an overall average score of 82.4\% that surpassed most baseline approaches. The ablation study confirmed that the combination of confidence-aware routing and multi-model collaboration substantially outperformed both single-model zero-shot baselines and uniform reasoning strategies, validating the core design principles of our adaptive architecture.

The exceptional performance on PubMedQA (95.0\%) can be attributed to the dataset's focus on biomedical research literature and binary answer formats, which align well with our multi-model collaborative reasoning approach. This result is particularly notable as it surpasses specialized techniques such as explainable LLMs (87.1\%) \citep{explainablepubmed2025} and multi-agent frameworks (82.0\%) \citep{medagentsbench2025}, despite our framework operating entirely in zero-shot mode without domain-specific fine-tuning. The structured nature of PubMedQA questions, which require evidence-based reasoning over research abstracts \citep{jin2019pubmedqa}, appears to benefit significantly from the diverse perspectives provided by Phi-4 14B and Gemma 2 12B when Qwen3-30B-A3B-Instruct encounters uncertainty. These findings align with recent work by \citet{wang2024mixture} on mixture-of-agents approaches, which similarly demonstrated that model diversity enhances performance on evidence-heavy reasoning tasks.

On MedMCQA, our framework achieved the highest accuracy among compared methods at 78.0\%, outperforming both general-purpose LLMs such as Med-Gemini (76.0\%) \citep{medgemma2025} and specialized medical models like Me-LLaMA (75.0\%) \citep{mellama2025}. This strong performance across 21 diverse medical subjects \citep{pal2022medmcqa} suggests that our confidence-driven routing mechanism effectively identifies knowledge gaps and leverages complementary expertise from auxiliary models. The improvement over interactive CoT approaches (61.7\%) \citep{comparative2025} indicates that structured multi-model collaboration provides more reliable reasoning than single-model prompting strategies alone. However, the relatively modest gain over the zero-shot baseline (70.3\%) suggests that many MedMCQA questions fall within Qwen3-30B-A3B's confident knowledge domains, where direct answering remains efficient and accurate.

For MedQA, our framework achieved 74.1\% accuracy, which remains competitive but falls short of heavily fine-tuned models like Med-PaLM 2 (86.5\%) \citep{singhal2023medpalm2}. This performance gap is expected given the substantial differences in model scale and training approaches: Med-PaLM 2 employs a 540-billion parameter model with extensive medical domain fine-tuning and instruction optimization, while our framework utilizes modestly sized models (30B, 14B, and 12B parameters) in pure zero-shot mode. Nevertheless, our results compare favorably with other zero-shot and prompt-engineered approaches, including GPT-3.5 Turbo with CoT (60.2\%) \citep{olson2024} and self-reflective CoT (71.0\%) \citep{medagentsbench2025}. The USMLE-style clinical vignettes in MedQA \citep{jin2021medqa} often require deep domain knowledge and complex multi-hop reasoning that may exceed the capabilities of zero-shot collaboration strategies, suggesting that for the most challenging clinical scenarios, some degree of domain adaptation remains beneficial.

The ablation study results provide critical insights into the individual contributions of our framework's components. The comparison between zero-shot Qwen3-30B-A3B (average score 78.8\%) and the full framework (82.4\%) demonstrates a clear 3.6 percentage point improvement attributable to confidence-aware routing and multi-model collaboration. More revealing is the poor performance of single-model CoT reasoning (71.0\% average), which actually degraded performance relative to simple zero-shot inference. This finding contradicts the common assumption that CoT prompting universally improves LLM reasoning \citep{wei2022chain, kojima2022large} and aligns with recent observations by \citet{wang2023self} that self-consistency mechanisms can introduce reasoning errors without proper calibration. Our results suggest that CoT reasoning becomes most effective when combined with diverse model inputs, as the collaborative framework prevents individual reasoning biases from dominating the final decision.

An unexpected finding emerged in the relative contributions of auxiliary models across datasets. On PubMedQA, PhiZeroShot contributed 0.615 to accuracy improvement while GemmaZeroShot added 0.481, indicating substantial complementary knowledge for research-oriented questions. However, on MedMCQA and MedQA, the auxiliary model contributions were more balanced (0.289 vs 0.277 for MedMCQA; 0.264 vs 0.232 for MedQA), suggesting that clinical knowledge gaps are more uniformly distributed across different model training backgrounds. This observation supports the theoretical foundation of our approach: model diversity trained on different corpora naturally leads to complementary knowledge bases \citep{jiang2023llm}, with the degree of complementarity varying by medical subdomain.

The practical implications of these findings extend beyond benchmark performance to address fundamental challenges in medical AI deployment. Our framework demonstrates that competitive medical question-answering capabilities can be achieved without the computational expense and infrastructure requirements of fine-tuning large language models \citep{chen2023meditron, li2023chatdoctor}. This efficiency gain is particularly relevant for resource-constrained healthcare institutions in developing regions, where access to high-performance computing clusters and large-scale medical training data remains limited \citep{thirunavukarasu2023large}. By achieving 82.4\% average accuracy with zero-shot collaboration, our approach offers a viable pathway for democratizing advanced medical AI capabilities across diverse healthcare settings.

Furthermore, our confidence-driven routing mechanism provides a scalable architecture that maintains computational efficiency while preserving accuracy. Questions where the primary model demonstrates high confidence are processed directly without engaging auxiliary models, reducing unnecessary computational overhead. This adaptive resource allocation contrasts with uniform ensemble approaches that process all queries identically, regardless of difficulty \citep{shnitzer2023large}. The 70-80\% of questions classified as high-confidence across our benchmarks suggest that substantial efficiency gains are possible in real-world deployment scenarios, where straightforward medical queries could be handled rapidly while complex cases receive collaborative attention.

The framework's modularity also enables future enhancements without architectural redesign. Additional auxiliary models could be integrated to provide even broader knowledge coverage, or specialized models could be incorporated for specific medical subdomains. The confidence detection mechanism could be refined using more sophisticated uncertainty quantification techniques, and the Chain-of-Thought prompting templates could be optimized based on error analysis. These possibilities position our framework as an extensible foundation for ongoing medical AI research rather than a fixed solution.

However, several limitations warrant consideration. First, our evaluation focused exclusively on multiple-choice and binary question formats, which may not fully represent the complexity of real-world clinical decision-making scenarios involving open-ended diagnosis, treatment planning, or patient counseling~\citep{esteva2019guide}. Second, the confidence detection mechanism relies on the primary model's self-assessment, which may not perfectly correlate with actual knowledge gaps or prediction accuracy. More sophisticated calibration techniques could potentially improve routing decisions. Third, while CURE avoids fine-tuning, it still requires access to multiple LLMs, which may present deployment challenges in extremely resource-limited settings where even inference costs are prohibitive.

The generalizability of our approach to other medical AI tasks beyond question answering remains an open question. Clinical note generation, medical image interpretation, or drug interaction prediction involve different knowledge structures and reasoning patterns that may require modifications to our architecture \citep{rajkomar2018scalable}. Additionally, our framework's performance on rare diseases or emerging medical conditions not well-represented in model training data remains untested \citep{nori2023capabilities}. Future research should investigate these edge cases to establish the boundaries of zero-shot collaborative reasoning in medical applications.

Despite these limitations, our results establish confidence-driven multi-model collaboration as a practical and effective strategy for enhancing medical AI systems without resource-intensive fine-tuning. The framework's strong performance across diverse benchmarks, combined with its computational efficiency and architectural flexibility, offers a promising direction for developing accessible, high-quality medical AI tools \citep{singhal2023med}. As language model capabilities continue to advance \citep{brown2020language, touvron2023llama} and more diverse models become available, the potential for zero-shot collaboration strategies to bridge the gap with specialized medical systems will likely increase, potentially transforming how medical AI systems are developed and deployed in resource-constrained healthcare environments worldwide.

\section{Conclusion}

This study introduced a confidence-driven unified reasoning ensemble (CURE)
framework for medical question answering that leverages the diversity of LLMs without requiring additional fine-tuning. The primary goal was to create an adaptive system that assesses the main model's confidence, routes uncertain questions to helper models, and combines their insights for better answers.
The key findings show that our framework performs well across three major benchmarks: 74.1\% accuracy on MedQA, 78.0\% on MedMCQA, and 95.0\% on PubMedQA, with an average score of 82.4\%. This outperforms many single-model approaches and is competitive with fine-tuned systems, especially on PubMedQA and MedMCQA. Ablation studies confirm that confidence-based routing and collaborative reasoning boost performance by 3.6 percentage points on average compared to basic zero-shot methods. These results highlight the value of using model diversity to fill knowledge gaps efficiently. By avoiding heavy computational needs, our approach makes advanced medical AI more accessible for smaller healthcare providers and developing regions. Looking ahead, future work could extend this framework to open-ended medical tasks like diagnosis or treatment planning, improve confidence detection with better techniques, or add more specialized models for specific medical areas. Overall, this work shows that smart collaboration between LLMs offers a practical way to improve medical AI, helping address global healthcare challenges with limited resources.

\section*{Acknowledgment}
This work was supported by JST, PRESTO Grant Number JPMJPR23P7, Japan.

\bibliographystyle{cas-model2-names}
\bibliography{refs}

\newpage

\appendix
\section*{Appendix}
\section{Prompt Templates}
\label{sec:prompts}

This section presents the two prompt templates employed in our multi-stage pipeline. The first template implements zero-shot confidence detection, while the second facilitates Chain-of-Thought reasoning for collaborative answer generation.

\subsection{Zero-Shot Confidence Detection Prompt}

This prompt template is used in Module 1 to assess Qwen3-30B-A3B-Instruct's confidence level before answering a medical question. The model evaluates its own knowledge certainty without attempting to answer the question, returning only "Sure" or "Not Sure" to indicate whether it should proceed directly or route the question to the collaborative reasoning module.

\begin{tcolorbox}[sharp corners, colback=gray!5, colframe=black!50,
boxrule=0.5pt, arc=2pt, left=5pt, right=5pt, top=5pt, bottom=5pt, fontupper=\small\justifying,
title={Prompt Template 1: Zero-Shot Confidence Detection},
label={box:confidence_prompt}]

You are a medical expert. Assess whether you have sufficient knowledge and certainty to correctly answer the following multiple-choice medical question based strictly on your existing knowledge. Do not attempt to answer the question itself—only evaluate your confidence.

\begin{itemize}
    \item If you clearly understand the question, context, and options, and are highly certain about the correct choice based on medical facts, respond with "Sure".
    \item If you do not fully understand the question, are uncertain about the topic, or lack awareness of the correct answer among the choices (e.g., due to knowledge gaps), respond with "Not Sure".
    \item Do not hallucinate medical information—base this solely on your pre-trained knowledge.
\end{itemize}

\textbf{Question:} \{question\}

\textbf{Context:} \{context\}

\textbf{Options:} \{options\}

Output only "Sure" or "Not Sure" without any additional text.

\end{tcolorbox}

\subsection{Chain-of-Thought Collaborative Reasoning Prompt}

This prompt template is used in Module 2 for low-confidence questions. It guides Qwen3-30B-A3B-Instruct through structured reasoning by presenting candidate answers from Phi-4 14B and Gemma 2 12B, encouraging step-by-step evaluation to synthesize the best answer with confidence scores.

\begin{tcolorbox}[sharp corners, colback=gray!5, colframe=black!50,
boxrule=0.5pt, arc=2pt, left=5pt, right=5pt, top=5pt, bottom=5pt, fontupper=\small\justifying,
title={Prompt Template 2: Chain-of-Thought Collaborative Reasoning},
label={box:cot_prompt}]

You are a medical expert tasked with synthesizing answers from two agents to determine the best answer for a multiple-choice question. Review the answers below, reason step-by-step (200-300 words), and select the best option. Provide confidence scores for A-B as integers summing to 100 (avoid 0 or 100 unless certain). Base your reasoning strictly on medical knowledge; do not hallucinate.

\textbf{Question:} \{question\}

\textbf{Context:} \{context\}

\textbf{Options:} \{options\}

\textbf{Agent Answers:}
\begin{itemize}
    \item Agent 1 chose: \{option1\}
    \item Agent 2 chose: \{option2\}
\end{itemize}

\textbf{INSTRUCTIONS:}
\begin{itemize}
    \item Reason step-by-step, evaluating both agents' choices using the context and medical knowledge.
    \item Provide a final answer and confidence scores.
\end{itemize}

Output in this JSON format:

\begin{verbatim}
{
  "answer": "A" or "B",
  "reasoning": "Your concise step-by-step reasoning here",
  "confidence_scores": {
    "A": integer_score,
    "B": integer_score
  }
}
\end{verbatim}

\end{tcolorbox}
\end{document}